\pdfoutput=1

\RequirePackage{etex}

\documentclass[runningheads]{llncs}
\usepackage[T2A]{fontenc}
\usepackage[utf8]{inputenc}
\usepackage[warn]{mathtext}
\usepackage{amsmath}
\usepackage{amsfonts}
\usepackage{mathrsfs}
\usepackage{mathtools}
\usepackage{csquotes}
\usepackage{graphicx}
\usepackage{newclude}
\usepackage{amssymb}
\usepackage{bbm}
\usepackage{float}
\usepackage{enumerate}
\usepackage{booktabs}
\usepackage{tabularx} 
\usepackage{tablefootnote}
\usepackage{xcolor}
\usepackage{arydshln}
\usepackage{hyperref}
\usepackage{multicol}
\usepackage{multirow}
\usepackage{mathbbol}
\usepackage{comment}
\usepackage{accents}
\usepackage[mode=buildnew]{standalone}
\usepackage{tikz}
\usepackage{tikz-network}
\usetikzlibrary{topaths,calc}
\graphicspath{{fig/}}
\DeclareGraphicsExtensions{.pdf,.png,.jpg}
\usepackage{color}

\begin{document}

\title{Formal concept analysis for evaluating intrinsic dimension of a natural language}
\author{Sergei O. Kuznetsov\inst{1}\orcidID{0000-0003-3284-9001} \and
Vasilii A. Gromov\inst{1}\orcidID{0000-0001-5891-6597} \and
Nikita S. Borodin\inst{1}\orcidID{0000-0002-7102-4443} \and
Andrei M. Divavin}

\titlerunning{FCA to evaluate intrinsic dimension of a natural language}
%
\authorrunning{S.O. Kuznetsov et al.}
%
\institute{HSE University, Moscow 109028, Russia\\}

\maketitle    
\begin{abstract}
Some results of a computational experiment for determining the intrinsic dimension of linguistic varieties for the Bengali and Russian languages are presented. At the same time, both sets of words and sets of bigrams in these languages were considered separately. The method used to solve this problem was based on formal concept analysis algorithms. It was found that the intrinsic dimensions of these languages are significantly less than the dimensions used in popular neural network models in natural language processing.
\end{abstract}
%
\section{Introduction}
The emergence of methods for representing words and n-grams
of a natural language as real-valued vectors (embeddings) allows one to ask about the definition of \emph{intrinsic
dimensions} of sets of words and bigrams observed in a given natural language. As a manifestation of the concept of intrinsic dimensions, let us consider  a sphere, which is a two-dimensional object, where every point is given by two coordinates - latitude and longitude.
Being embedded in three-, five- and ten-dimensional space, it will
be given by a set of vectors, respectively, with three, five and ten
coordinates, however, but it will remain a two-dimensional object, and
its intrinsic dimension will be equal to two. The problem of intrinsic dimension
is important  from the practical point of view: its solution will allow one to judge the appropriateness of using  very large vectors of embeddings (first of all, neural network models like BERT, etc.) in NLP domain.

\section{Related work}
Turning to methods for estimating the intrinsic dimension, we note, first of all, the
work of V.~Pestov on intrinsic dimension of a dataset \cite{pestov-2007}, where the requirements for the definition of intrinsic dimension and methods for obtaining it were formulated.
The introduced axiomatics is based on the results of M. Gromov~\cite{m_gromov}.

It seems to us that the approaches known from the literature can be divided
into three large classes:  methods using 
a method for estimating the dimension of the strange attractor \cite{kantz_schreiber}; graph-based methods ~\cite{costa_et_al, farahmand_et_al, kozma_et_al}; methods based
to one or another variant of persistent homology analysis ~\cite{schweinhart}. Let us take a look at the three above
approaches.

First of all, among the works related to the establishment of the dimensions of the strange attractor
 we note a classical monograph by Kantz and Schreider~\cite{kantz_schreiber}. It considers a set of classical approaches to the definition of the concept
dimension of a strange attractor (topological, Hausdorff, spectrum,
generalized entropy dimensions, etc.) and methods for their evaluation.
Unfortunately, the classical approaches to determining the dimension of the strange
attractors suffer from two
disadvantages: firstly, they usually require very significant
computing power and, secondly, are often non-robust with respect to
to sample changes. These circumstances necessitated
creation of a new generation of methods for estimating the intrinsic dimension.

Kozma et al.~\cite{kozma_et_al} proposed a method of estimating the upper box
dimension using a minimum spanning tree and
statistics based on it. J.A. Costa et al. ~\cite{costa_et_al} and A. Farahmand et al. ~\cite{farahmand_et_al} in numerous papers have developed the idea of estimating intrinsic dimension by examining nearest neighbor graphs. In 2013 M.R. Brito et al. ~\cite{brito_et_al} proposed an approach based on nearest neighbor graph (KNN), minimum weight spanning tree (MST) and sphere of influence (SOI) analysis to determine the Euclidean dimension of i.i.d. geometric dataset.

Adams et al.~\cite{adams_et_al} suggested a way to computing
intrinsic dimension using persistent homology
(Persistent homology dimension). The idea is to investigate the properties of random variables of the following form:
\begin{equation}
    E_\alpha^{i}(x_1,\ldots,x_n) = \sum\limits_{I\in {PH}_i(x_1,\ldots,x_n)}\left|I\right|^\alpha,
\end{equation}
where $\{x_j\}_{j\in\mathbb{N}}$ are i.i.d. samples from a probability measure on a metric space, ${PH}_i(x_1,\ldots,x_n)$ denotes the i-dimensional reduced persistent homology of the Čech
or Vietoris–Rips complex of $\{x_1,\ldots,x_n\}$, and $\left|I\right|$ is the length of a persistent homology
interval. Schweinhart~\cite{schweinhart} carried out a rigorous formal analysis of this
estimates, its connection is established with the upper box dimension.  Jaquette and Schweinhart extended the methodology  to the case of fractal systems,
characterized by non-integer dimensions.

We also note a  stand-alone work~\cite{hanika_et_al}, based on
the concept of distance between metric spaces, introduced by M.Gromov~\cite{m_gromov} and formal concept analysis ~\cite{ganter_kuznetsov, kuznetsov}.


\section{Main definitions and problem statement}

\subsection{Formal Concept Analysis and Pattern Structures}

Here we give basic definitions and facts related to Formal Concept Analysis~\cite{kuznetsov} and Pattern structures from~\cite{ganter_kuznetsov}.

In basic FCA  binary data are given by a \emph{formal context} $K = (G, M, I)$, where $G$ is the set of objects, $M$ is the set of (binary) attributes, and $I$ is a incidence relation between objects and attributes: $I \subseteq G \times M$.

\emph{Derivation (prime) operators} $(\cdot)'$ are defined as follows:  $A'$  gives the subset of all attributes shared by all objects from $A \subseteq G$. Similarly, $B'$ gives the  subset of all objects having all attributes from $B \subseteq M$.
\begin{align}
    A' &= \{m \in M \mid \forall g \in A: gIm \}, \\
    B' &= \{g \in G \mid \forall m \in B: gIm \}, 
\end{align}

A \emph{formal concept} is a pair $(A,B)$ of subsets of objects $A$ and attributes $B$, such that  $A=B'$, and $B=A'$.

The generality order $\leq$ on concepts is defined as follows:
$$ (A_1,B_1) \leq (A_2,B_2) \mbox{if} A_1\subseteq A_2 (\Leftrightarrow B_2\subseteq B_1).$$

The set of all concepts makes an  algebraic lattice, called \emph{concept lattice}, w.r.t. the partial order $\leq$, so that every two concepts have supremum and infimum w.r.t. $\leq$.

Pattern structures~\cite{ganter_kuznetsov, kuznetsov} propose a generalization of FCA so that objects, instead of sets of binary attributes, can be described by complex descriptions, the set of which is ordered w.r.t. subsumption (containment) relation.

Let $G$ be a set of objects, let $(D, \sqcap)$ be a meet-semilattice of  \emph{descriptions}, and let  $\delta:G \rightarrow D$ be a mapping that assigns a description to each object from $G$. Then $(G,\underline{D},\delta)$, where $\underline{D} = (D,\sqcap)$; is called a \emph{pattern structure} w.r.t. \enquote{similarity operation} $\sqcap$, provided that the set $\delta(G) \coloneqq \{\delta(g)\:|\:g\in G\}$ generates a complete subsemilattice $(D_{\delta},\sqcap)$ of $(D,\sqcap)$, i.e., every subset $X$ of $\delta(G)$ has an infimum $\sqcap X$ in $(D,\sqcap)$ and $D_{\delta}$ is the set of these infima.

If $(G,\underline{D},\delta)$ is a pattern structure, the \emph{derivation (prime) operators} are defined as

    \[A^{\diamond} \coloneqq \underset{g \in A}{\sqcap} \delta(g) \enspace \mbox{for all} \enspace A \subseteq G \] 
    
   
    \[d^{\diamond} \coloneqq \{ g \in G \, \mid \, d \sqsubseteq \delta(g)\}\enspace \mbox{for all} \enspace d \in D  \] 
    
    The set $D$ is partially ordered w.r.t. the following \emph{subsumption} relation:
    
    \[ c \sqsubseteq d \colon \Longleftrightarrow c \sqcap d = c \] 
    
  A \emph{pattern concept} of $(G,\underline{D},\delta)$ is a pair $(A,d)$ satisfying 
    
    \[ A \subseteq G, \, d \in D, \, A^{\diamond} = d \enspace \mbox{and} \enspace A = d^{\diamond}\] 
    
    The set of all pattern concepts forms the \emph{pattern concept lattice}. 
    
    In this paper we will use interval pattern structure~\cite{kknd11, kuznetsov}, an important case of pattern structures where descriptions from $D$ are tuples of closed numerical intervals of the form $[s,t]$, $s,t \in \mathbb{N}$ and the similarity operation on two interval tuples is given  component-wise as their convex hull:
    $$ [s,t] \sqcap [q,r] = [\mbox{min}\{s,q\}, \mbox{max}\{t,r\}].$$

    The intuition of this definition is explained by interpreting numerical intervals as uncertainty intervals, so that their similarity is a minimal convex cover of both uncertainties.

\subsection{Intrinsic Data Dimension and Problem Statement}

In Hanika et al.~\cite{hanika_et_al}, the authors suggest using approaches based on Gromov’s metric~\cite{m_gromov} and Pestov’s axiomatic approach~\cite{pestov-2008} to estimate the intrinsic dimension of
data. So, given tabular data in the form of a formal context
$\mathbb{K} = (G,M,I)$ and measures $\nu_G$, $\nu_M$ on sets $G$ and $M$ of objects and attributes, respectively, the complexity of the dataset (called \emph{observed diameter}) is defined as

\begin{align}
Diam(\mathbb{K}, \alpha)=\mbox{max}\{\nu_G(A)\mid(A,B)\in\mathfrak{B}(\mathbb{K}), \alpha<\nu_M(B)<1-\alpha\},
\end{align}

where $\alpha$ is a parameter.
If datatable is not binary, but e.g., given by an interval pattern structure $\mathbb{PS}$, then this definition can be extended as follows:

\begin{align}
Diam(\mathbb{PS}), \alpha)=\mbox{max}\{\nu_G(A)\mid (A,d)\in\mathfrak{B}(\mathbb{PS}), \alpha<\nu_M(d)<1-\alpha\}, 
\end{align}
where $\alpha$ is a parameter.
A natural specification of the general form of $\nu_M$ in this definition would be 
$$ \sum_{i\in M} \nu_i \cdot  \mathbb{1}_{(y_i - x_i) \leq \theta},$$
where $[x_i, y_i]$ is the $i$th interval component of tuple $d$, $\nu_i$ is the measure (weight) associated to attribute $i\in M$, and $\theta$ is a parameter.

The intrinsic dimension of the context (dataset) is then defined as follows:
\begin{align}
Dim(\mathbb{K})=\left(\int_{0}^{1}Diam\left(\mathbb{K},\alpha\right)\delta\alpha\right)^{-2},
\end{align}

Now the problem we solve in this paper can formally be stated as follows.

Given a set of natural language texts $C=\{T_1, \ldots, T_N\}$; $d\in \mathbb{N}$, the dimension of the embedding space; $n\in \mathbb{N}$, the number of words in $n$-grams
   (it is assumed that the set of texts $C$ is a representative sample of texts of the corresponding natural language)
   
\begin{enumerate}

\item
compute the sets $E_n(d), n\in\mathbb{N}$ of
embeddings of texts from $C$, construct the set $F$ of respective threshold binary attributes $F$, where $f(e)\in F$ means $e>0$ for $e\in E_n(d)$, 

\item
compose the context $\mathbb{K} = (C, F, I)$, with the relation 
$$I = \{(T,f(e)) \mid e(T) > 0 \mbox{ for embedding } e \mbox{ of text } T \},$$  

\item
calculate the approximate value of (6) by using the trapezoid method for integration:
\begin{align}
Dim(\mathbb{K}; \ell)=\left(\frac{1}{2\ell}\sum_{i=1}^{\ell}\left[Diam\left(\mathbb{K},\frac{i-1}{\ell}\right)+Diam\left(\mathbb{K},\frac{i}{\ell}\right)\right]\right)^{-2},
\end{align}

here $\ell$ is the number of intervals used for approximation.

\end{enumerate}

\section{Realization of the model}

\subsection{Computing Data Dimension}

To compute data dimension according to (7) one does not need to compute the set of all concepts, which can be exponentially large w.r.t. initial data. One can use the properties of monotonicity of measures $\nu_G$ and $\nu_M$ and antimonotonicity of prime operators $(\cdot)'$.
The general idea of computing the observed diameter is as follows: start with largest possible concept extents (corresponding to one-element sets of attributes), which have largest $\nu_G$ measure, and decrease them until the measure $\nu_M$ of respective concept intents falls in the interval $[\alpha; 1 - \alpha]$.

More formally: 

\begin{itemize}

\item[1.] For every attribute  $m \in M$ compute $m'$ and $m''$, generating concepts  $(A,B)$, where $A = m'$, and $B = m''$. 

\item[2.] If some of the generated concepts  satisfy (4), take concept $(A,B)$ with the largest $\nu_G(A)$ as the solution to the problem.

\item[3.] If there are no $(A,B)$ satisfying (4), for every $(A,B)$ generated so far compute $((B \cup \{n\})', (B \cup \{n\})''$, where $n\not\in B$.

\item[4.] Iterate steps 1, 2 and 3 until solution is obtained for all given $\alpha_i\in\left\{\alpha\right\}_\ell\subseteq [0;1]$. 

\item[5.] Compute $Dim(\mathbb{K; \ell})$. 

\end{itemize}

Since the cardinality of intents are growing with every iteration of Steps 2 and 3, finally the process will attain intents satisfying $\alpha < \nu_M(B) < 1 - \alpha$. It is no need to go further by increasing intents, because with increasing intents, extents will decrease with no more chance to obtain a concept extent with the largest measure $\nu_G$ satisfying (4).

What is the computational complexity of first four steps for a single value of $\alpha$? If $\alpha = 0$, then, by monotonicity of measure $\nu_G$, the algorithm terminates at the first iteration of step 2 by outputting max~$\nu_G(m')$ for $m\in M$, which takes $O(|M|)$ applications of prime operation $(\cdot)'$. If $\alpha\neq 0$, then one needs to generate intents with $\nu_M \leq \alpha$ until the subsequent generation in step 3 would produce intents with  $\nu_M > \alpha$, 
so that the observed diameter would be the maximal $\nu_G$ of respective extents.
For $\alpha>0$ let $k(\alpha)$ denote the number of concepts $(A,B)$ with $\alpha < \nu_M(B) < 1 - \alpha$. Then, applying the argument of ``canonicity check''~\cite{kuz93}, one can show that the algorithm terminates upon $O(|M|\cdot k(\alpha))$ applications of $(\cdot)'$. 

\subsection{Preprocessing text corpora}

To compute language dimensions we take standard open source Internet language corpora that represent national prose and poetry for Bengali, English and Russian languages.
The preprocessing of corpora consists
of several steps. 
\begin{enumerate}[1{)}]
\item Removal of stop words: some words are found in large numbers in texts that affect a variety of subject areas and, often, are not in any way informative or contextual, like articles, conjunctions, interjections, introductory words;
\item Tokenization: other words can provide useful information only by their presence as a representative of a certain class, without specificity; good examples for tokenization are proper names, sentence separators, numerals;
\item Lemmatization: it is useful to reduce the variability of language units by reducing words to the initial form
\end{enumerate}
Then we apply standard techniques based on tf-idf measure to select keyterms from obtained $n$-grams of the texts, so that every text in the corpus is converted to a set of keyterms and the text-keyterm matrix is generated. Upon this we apply SVD-decomposition of this matrix and select first high-weight components of the decomposition to obtain semantic vector-space of important features of the language and allow for computing respective embeddings of the texts from the corpus. This results in obtaining text-features matrix, which is originally numerical. Then the numerical matrix can be either converted to binary one by selecting  value thresholds or treated directly by means of interval pattern structures described above.

\section{Computer Experiments}

We used the approach described above to estimate the intrinsic dimension of Russian, Bengali, and English languages. To do this,  we form  groups of sample sets of $n$-grams for different parameters $n$ and $d$. Then, for each $n\in{1,2}$ (103952 and 14775439 Russian words and bigrams; 209108 and 13080621 for Bengali words and bigrams; 94087 and 9490603 English words and bigrams), we average the obtained intrinsic dimensions over $d$ and round it to the nearest integer. We took the most obvious realization of the measures as $\nu_M(X) = |X|$ and $\nu_G(Y) = |Y|$. The results are presented in the Tables~(\ref{results_table_n1}, \ref{results_table_n2}) for 1,2-grams, respectively.

\vspace{-10pt}
\begin{table}
\caption{Intrinsic dimension estimation for natural languages, n=1.}\label{results_table_n1}
\begin{tabularx}{\textwidth} { 
  | >{\raggedright\arraybackslash}X 
  | >{\centering\arraybackslash}X
  | >{\centering\arraybackslash}X 
  | >{\centering\arraybackslash}X|}
\hline
Language & $d$ & Intrinsic dimension\\
\hline
\multirow{4}{*}{Bengali} & 5 & 6.2509\\ \cline{2-4} 
& 8 & 5.2254\\ \cline{2-4} 
& 14 & 4.7445\\ \cline{2-4} 
& 20 & 4.5318\\
\hline
\multirow{4}{*}{Russian} & 5 & 6.2505\\ \cline{2-4} 
& 8 & 5.2308\\ \cline{2-4} 
& 14 & 4.6470\\ \cline{2-4} 
& 20 & 4.4438\\
\hline
\multirow{4}{*}{English} & 5 & 6.2509\\ \cline{2-4}
& 8 & 5.2302\\ \cline{2-4} 
& 14 & 4.6437\\ \cline{2-4} 
& 20 & 4.4369\\
\hline
\end{tabularx}
\end{table}

\begin{table}
\caption{Intrinsic dimension estimation for natural languages, n=2.}\label{results_table_n2}
\begin{tabularx}{\textwidth} { 
  | >{\raggedright\arraybackslash}X 
  | >{\centering\arraybackslash}X
  | >{\centering\arraybackslash}X 
  | >{\centering\arraybackslash}X |}
\hline
Language & $d$ & Intrinsic dimension\\
\hline
\multirow{2}{*}{Bengali} & 5 & 5.6792\\ \cline{2-4} 
& 8 & 5.6346\\ \cline{2-4} 
\hline
\multirow{2}{*}{Russian} & 5 & 5.8327 \\ \cline{2-4} 
& 8 & 5.6610 \\ \cline{2-4} 
\hline
\multirow{2}{*}{English} & 5 & 5.4013 \\ \cline{2-4} 
& 8 & 5.6058 \\ \cline{2-4} 
\hline
\end{tabularx}
\end{table}

\section{Conclusion and future directions}
We have applied the combination of Formal Concept Analysis~\cite{kuznetsov} with an approach based on M. Gromov metrics~\cite{m_gromov, hanika_et_al} to estimating intrinsic  dimensions of linguistic structures with complex statistical characteristics. It is striking that for natural languages we observe very small values of intrinsic dimensions.

Namely,  the indicated dimension was found to be
  $\approx 5$ for all given languages. This fact allows us to conclude that the orders of intrinsic dimension are equal for the above languages.
  
Obvious directions for future research can be
analysis of other definitions of language dimension;
extending the list of languages for similar analysis, including languages from
various language families.

\section*{Acknowledgements}

The work of Sergei O. Kuznetsov was supported by the
Russian Science Foundation under grant 22-11-00323 and performed at HSE
University, Moscow, Russia.



\end{document}